\documentclass[a4paper]{article}
\pdfoutput=1 
\usepackage{INTERSPEECH2021}
\usepackage{graphicx,multirow,soul,boldline,hyperref,subcaption,xcolor,soul,flushend}

\title{Leveraging Pre-trained Language Model for Speech Sentiment Analysis}
\name{Suwon Shon$^{*1}$, Pablo Brusco$^{*1}$, Jing Pan$^1$, Kyu J. Han$^1$, Shinji Watanabe$^2$}
\address{
  $^1$ASAPP, USA\\
  $^2$Carnegie Mellon University, USA}
\email{\{sshon,pbrusco,jpan,khan\}@asapp.com, shinjiw@ieee.org}

\vspace{-0.3cm}

\begin{document}
\maketitle
\begin{abstract}

In this paper, we explore the use of pre-trained language models to learn sentiment information of written texts for speech sentiment analysis.
First, we investigate how useful a pre-trained language model would be in a 2-step pipeline approach employing Automatic Speech Recognition (ASR) and transcripts-based sentiment analysis separately.
Second, we propose a pseudo label-based semi-supervised training strategy using a language model on an end-to-end speech sentiment approach to take advantage of a large, but unlabeled speech dataset for training.
Although spoken and written texts have different linguistic characteristics, they can complement each other in understanding sentiment.
Therefore, the proposed system can not only model acoustic characteristics to bear sentiment-specific information in speech signals, but learn latent information to carry sentiments in the text representation.
In these experiments, we demonstrate the proposed approaches improve F1 scores consistently compared to systems without a language model.
Moreover, we also show that the proposed framework can reduce 65\% of human supervision by leveraging a large amount of data without human sentiment annotation and boost performance in a low-resource condition where the human sentiment annotation is not available enough.

\end{abstract}
\noindent\textbf{Index Terms}: speech sentiment analysis, pre-trained language model, end-to-end speech recognition

\section{Introduction}
{\let\thefootnote\relax\footnotetext{*Equal contribution}}

Speech sentiment analysis is the task of classifying positive/neutral/negative sentiments of a given speech.
Compared to emotion recognition, it is a more abstract level of a recognition task.
For example, negative sentiment not only contains anger emotion, but it also includes disparagement, sarcasm, doubt, suspicion, frustration, etc~\cite{mohammad2016practical}.
These negative sentiments may be related to the acoustic/prosodic features of speech as well as relevant to the context of the speech.

The conventional approach for speech sentiment analysis is using ASR on speech then employing sentiment analysis on the ASR transcripts so that it becomes a text classification task in a 2-step pipeline or cascade pipeline. 
However, this 2-step approach has two major disadvantages. First, it loses rich acoustic/prosodic information which is critical to understand spoken language. Second, there is a lack of sentiment-annotated datasets available when it comes to the spoken conversations domain, therefore, systems trained on a different communication channel from the conversational speech would tend to under-perform in production environments in the wild~\cite{lu2020speech}.
To address the first drawback, recently end-to-end (E2E) type speech sentiment analysis systems were proposed~\cite{lu2020speech,li2019dilated,li2018attention,wu2019speech,xie2019speech,tzirakis2018end,mirsamadi2017automatic}. For the second problem, one option can be collecting more sentiment-labeled spoken data such as \cite{chen2020large}, but such options would require a costly effort. 

\begin{figure}[t]
     \centering
     \begin{subfigure}[b]{0.8\linewidth}
         \centering
         \includegraphics[width=\textwidth]{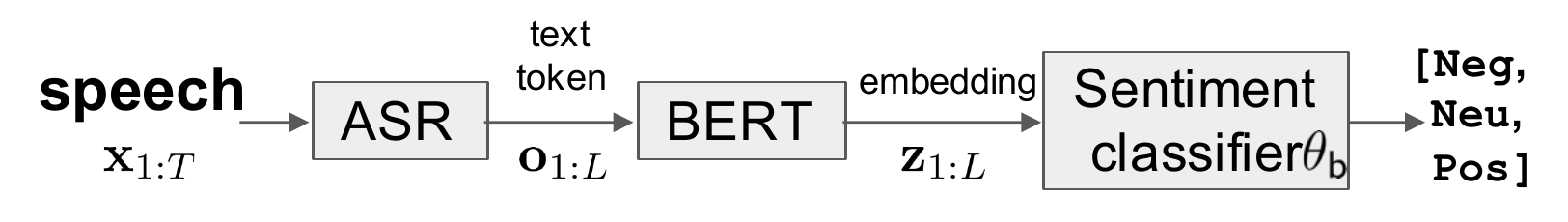}
         \vspace{-0.5cm}
         \caption{2-step pipeline}
     \end{subfigure}
     \begin{subfigure}[b]{0.85\linewidth}
         \centering
         \includegraphics[width=\textwidth]{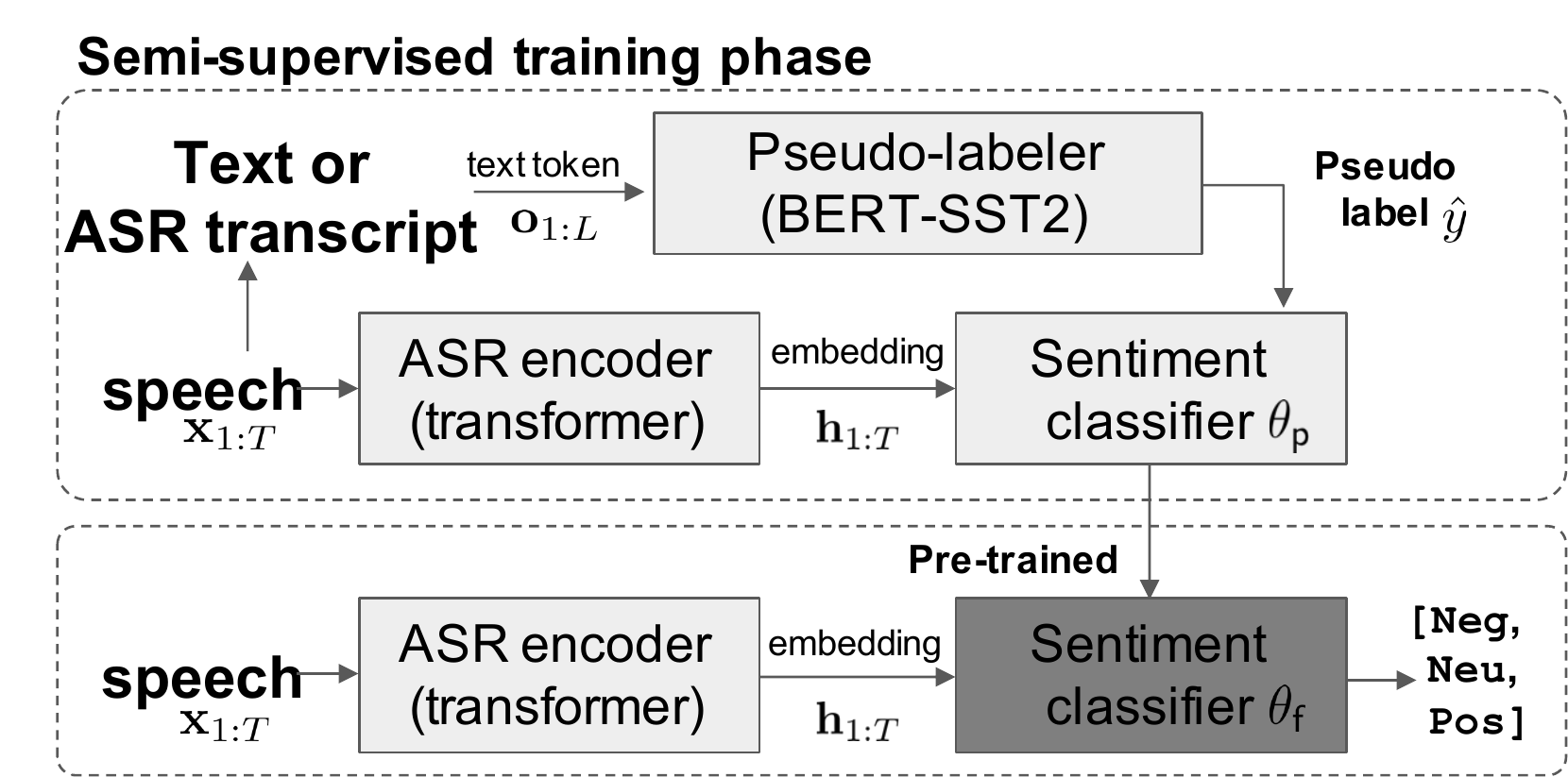}
         \vspace{-0.5cm}
         \caption{E2E system with semi-supervised training}
     \end{subfigure}
     \vspace{-0.3cm}
    \caption{Proposed speech sentiment analysis}
    \label{fig:concept}
    \vspace{-0.6cm}
\end{figure}

In this paper, to tackle these two problems all together, we investigate how we can minimize human supervision and, at the same time, still efficiently train a model for speech sentiment analysis. In the Natural Language Processing (NLP) field, we can easily obtain robust sentiment analysis models by fine-tuning powerful pre-trained LMs such as BERT \cite{devlin2018bert}. 
We thus leverage the pre-trained LMs to provide the additional information obtained through large training datasets for speech sentiment analysis models.
We evaluate this approach in a 2-step pipeline setup first and compare it against an E2E framework in learning speech sentiments with limited training data. 
For the 2-step setup, we use a pre-trained LM as is, then process ASR transcripts as input for text-based sentiment analysis as shown in Fig.\ref{fig:concept}(a).
For the E2E model, we use a semi-supervised training approach using LM as a pseudo labeler as shown in Fig.~\ref{fig:concept} (b). 
In this system, we use an ASR encoder and pseudo labeler to train a sentiment classification model semi-supervisedly with pseudo labeled data then further tune the sentiment classification model with smaller, annotated data in the fine-tuning stage. We compare our system with the state-of-the-art system proposed in~\cite{lu2020speech}. 
The evaluation was performed on a large sentiment dataset published by~\cite{chen2020large} 
that mitigates drawbacks in conventional datasets such as being scripted~\cite{busso2008iemocap} or single speaker monologues~\cite{zadeh2018multimodal}.

\section{Related work}
Learning good representation from speech signals is the key to a speech sentiment/emotion analysis task. 
A recent study suggests to use a pre-trained ASR encoder~\cite{lu2020speech} to prevent overfitting, and it showed promising results by surpassing the traditional audio + text multimodal systems~\cite{kim2019dnn, siriwardhana2020jointly,cho2018deep}.
Without the pre-trained ASR encoder, the model tends to overfit to the training data and the same model working on emotion recognition gives mediocre results on the sentiment analysis task~\cite{lu2020speech}.

Similarly, in the study of Spoken Language Understanding (SLU), pre-training approaches were proposed in combination with ASR~\cite{kuo2020end,chen2018spoken,haghani2018audio} or acoustic classification modules~\cite{lugosch2019speech}, using ground truth (GT) text or ASR transcripts to improve SLU performance under limited resources.

The aforementioned pre-training approaches are based on the assumption that if a model is pre-trained to recognize words or phonemes, the fine-tuning result of downstream tasks will be improved.
Our approach is also based on the same assumption, but we propose the use of powerful pre-trained LMs to transfer more abstract knowledge from the written text-domain to speech sentiment analysis.
Specifically, we leverage pre-trained BERT models to extract robust embedding from text tokens for the 2-step pipeline, and to generate pseudo labels for semi-supervised training a model for the E2E speech sentiment analysis system.

\section{Approaches}

In the field of NLP, great advances have been made through pre-training task-agnostic LMs without any supervision. 
These pre-trained models can be fine-tuned using downstream task-specific data and showed state-of-the-art performance in many problems such as text classification, question answering, and summarization~\cite{devlin2018bert,sanh2019distilbert,yang2019xlnet,liu2019roberta}.

In this section, we describe how we use LMs, such as BERT, for speech sentiment analysis.
First, we place the pre-trained LM as embedding layers for the sentiment classification model in the 2-step pipeline framework.
Second, we explore how the speech sentiment classifier in the E2E system can be enhanced through the use of a pseudo label-based semi-supervised training approach applied to large audio corpora without human annotations.
By leveraging the LM in both the 2-step pipeline and E2E framework, we 
expect models to generalize better by being able to integrate the text-domain sentiment-related knowledge into the speech sentiment analysis space.

\vspace{-1mm}
\subsection{2-step pipeline}
\label{sec:2step:describe}
\vspace{-1mm}

Here we describe how we build two systems that use ASR transcripts as input: a baseline system in which embeddings are trained from scratch, and a BERT-based model that uses the pre-trained BERT as an embedding layer.

Suppose an input acoustic feature sequence is $\textbf{x}_{1:T}$ and an ASR transcript or GT transcript token sequence is $\textbf{o}_{1:L}$. Then, for the baseline system, the token sequence $\textbf{o}_{1:L}$ is followed by BLSTM layers and an output layer to predict the three possible sentiments, e.g. Negative, Neutral, and Positive. Therefore, the model is trained by maximizing
$P(y|\theta_{\mathsf{s}},\textbf{o}_{1:L}) $
where $\theta$ represents the model parameters and $y$ the GT sentiment label.
The BERT-based system uses a pre-trained BERT model (bert-base-uncased~\cite{wolf2019huggingface}) to encode the token sequence $\textbf{o}_{1:L}$ into a BERT output sequence $\textbf{z}_{1:L}$, after which the same sentiment classification layers follow as the baseline system has. The function to maximize in this case is
$P(y|\theta_{\mathsf{b}},\textbf{z}_{1:L})$.
In Section \ref{sec:2step:setup} we describe more details about these two architectures.

\begin{table*}[t]
\centering
\caption{2-step pipeline evaluation on pre-trained BERT systems vs. the baseline systems. All the models were evaluated on ASR transcripts (SWBD-test/holdout-ASR means ASR transcripts). If not specified, the models use the full SWBD-train set (86h).}
\vspace{-0.3cm}
\label{tab:2step_performnance_bert}
\resizebox{0.92\textwidth}{!}{%
\begin{tabular}{c|c|ccc|ccc|ccc|ccc}
\hlineB{2}
 & SWBD-train & \multicolumn{6}{c|}{Validation Set (SWBD-test-ASR)} & \multicolumn{6}{c}{Evaluation Set (SWBD-holdout-ASR)} \\ \cline{3-14}
 & transcript & \multicolumn{3}{c|}{Unweighted} & \multicolumn{3}{c|}{Weighted} & \multicolumn{3}{c|}{Unweighted} & \multicolumn{3}{c}{Weighted} \\ \cline{3-14}
 Architecture &  type & REC & PRE & F1 & REC & PRE & F1 & REC & PRE & F1 & REC & PRE & F1 \\ \hlineB{2}
\multirow{3}{*}{Baseline} & GT & 59.27 & 55.55 & 55.06 & 56.39 & 62.88 & 57.50 & 59.21 & 55.77 & 55.55 & 56.93 & 62.10 & 57.91 \\ \cline{2-14}
& ASR & 52.57 & 50.07 & 47.38 & 47.60 & 58.44 & 48.30 & 52.43 & 49.84 & 47.66 & 47.92 & 57.12 & 48.48 \\ \cline{2-14}
& ASR (5h) & 33.33 & 18.31 & 23.64 & 54.93 & 30.18 & 38.96 & 33.33 & 17.82 & 23.22 & 53.46 & 28.58 & 37.24 \\ \hline

\multirow{3}{*}{BERT}     & GT & \textbf{63.87} & 64.64 & \textbf{64.12} & \textbf{68.16} & 68.01 & \textbf{67.96} & \textbf{64.53} & 65.05 & \textbf{64.56} & \textbf{67.87} & \textbf{67.98} & \textbf{67.73} \\ \cline{2-14}
& ASR & 63.75 & \textbf{65.21} & 63.87 & 68.08 & \textbf{68.46} & 67.78 & 63.63 & \textbf{65.13} & 63.64 & 67.29 & 67.94 & 66.99 \\ \cline{2-14}
& ASR (5h) & 50.18 & 55.01 & 50.99 & 61.08 & 58.88 & 58.82 & 50.09 & 56.06 & 51.03 & 60.85 & 58.91 & 58.34 \\ \hlineB{2}

\end{tabular}
}
\vspace{-0.15cm}
\end{table*}

\begin{table*}[ht]
\centering
\caption{E2E speech sentiment analysis baseline evaluation. SWBD-train set was used for training.}
\vspace{-0.3cm}
\label{tab:e2e_baseline}
\resizebox{0.92\textwidth}{!}{%
\begin{tabular}{c|c|ccc|ccc|ccc|ccc}
\hlineB{2}
            & Sentiment & \multicolumn{6}{c|}{Validation Set (SWBD-test)} & \multicolumn{6}{c}{Evaluation Set (SWBD-holdout)} \\ \cline{3-14} 
              &        Classifier      & \multicolumn{3}{c|}{Unweighted} & \multicolumn{3}{c|}{Weighted} & \multicolumn{3}{c|}{Unweighted} & \multicolumn{3}{c}{Weighted} \\ \cline{3-14} 
Input feature & Architecture & REC    & PRE    & F1   & REC   & PRE   & F1   & REC    & PRE    & F1   & REC   & PRE   & F1   \\ \hlineB{2}
FBank     & CNN & 41.94&	47.88&	41.87&	56.21&	52.94&	51.73&	40.00&	45.62&	38.90&	51.68&	49.68&	46.70    \\ \hline
RNN-T encoder~\cite{lu2020speech} & BLSTM & 62.39    &  -   &  -  &  70.10  &-    &-    &    - & -    & -    &  -   &    - &  -  \\ \hline
CTC-Attention encoder & BLSTM & 64.59&	68.89&	66.24&	71.41&	70.86&	70.72&	61.21&	65.92&	62.74&	67.73&	67.89&	66.99    \\ \hlineB{2} 
\end{tabular}%
}
\vspace{-0.3cm}
\end{table*}

\vspace{-1mm}
\subsection{Semi-supervised E2E speech sentiment analysis}
\vspace{-1mm}
As for the E2E systems, we start by creating a baseline system that uses an ASR encoder output as features for speech sentiment analysis~\cite{lu2020speech}. 
Based on this framework, we propose a pseudo label-based semi-supervised training approach, shown in Figure~\ref{fig:concept}.(b), that we describe in the next subsections.

\subsubsection{Sentiment classifier}
Let $\textbf{h}_{1:T}$ be the ASR encoder output given $\textbf{x}_{1:T}$. The sentiment classifier block takes $\textbf{h}_{1:T}$ as input and predicts a sentiment class. We follow a similar architecture to the one proposed in ~\cite{mirsamadi2017automatic}. 
This architecture has two BLSTM layers after a fully connected (FC) layer. Then, an attention-based weighted pooling takes the output sequence of the BLSTM, so that the average is performed to summarize the frame-level embedding into an utterance level embedding. Finally, the output layer maps the utterance-level embedding into a sentiment class. As the cost function, we used a cross entropy loss.

\subsubsection{Semi-supervised training with pseudo label}

To transfer the knowledge from the text domain, we generated sentiment pseudo labels $\hat{y}$ using a pre-trained LM, \textit{a pseudo labeler} from the given token sequence $\textbf{o}_{1:L}$, which can be generated either from the GT or the ASR transcripts. Then, we use the pseudo labels to train the sentiment classifier.

For building this pseudo labeler, we first chose a few state-of-the-art pre-trained LMs, i.e. BERT~\cite{devlin2018bert}, DistilBERT~\cite{sanh2019distilbert}, RoBERTa~\cite{liu2019roberta}, XLNet~\cite{yang2019xlnet}, that we fine-tuned with the Stanford Sentiment Treebank (SST) data \cite{socher-etal-2013-recursive}. At the end of this process, we obtained four different text-based sentiment analysis models that we will use as pseudo labelers in our experiments.
The semi-supervised training of the sentiment classifier using pseudo labels can be done by maximizing $P(\hat{y}|\theta_{\mathsf{p}},\textbf{h}_{1:T})$ to pre-train $\hat{\theta}_{\mathsf{p}}$, and fine-tuning by maximizing
$P(y|\theta_{\mathsf{f}},\textbf{h}_{1:T};\hat{\theta}_{\mathsf{p}})$.

\section{Experiments}
\subsection{Datasets and Metric}
For our experiments, we used the SWBD-Sentiment dataset \cite{chen2020large}
labeled with 3 sentiments (negative, neutral, and positive) by 3 different human annotators for every segment. From each segment, we computed the majority vote and discarded utterances in which there was a 3-way disagreement. 
We split the resulting data into a 86h training set (SWBD-train), a 5h test set (SWBD-test), and a 5h holdout set (SWBD-holdout)\footnote{The split information was provided by the authors of \cite{chen2020large}.}. We use SWBD-test as our validation set during training for choosing the best hyperparameters. We used SWBD-holdout as our evaluation dataset. 

During evaluation, we computed weighted and unweighted averages of recall (REC), precision (PRE), and F1 scores (F1). Note that the conventional weighted/unweighted accuracy is equivalent to the weighted/unweighted REC since speech sentiment analysis is a closed-set multi-class classification task.

\subsection{2-step pipeline experiment setup}
\label{sec:2step:setup}
In some of our experiments, instead of using the GT texts, we used the ASR transcripts of the SWBD-Sentiment and Fisher datasets~\cite{cieri2004fisher}. These were generated by using an HMM-DNN hybrid ASR model with a multistream CNN architecture~\cite{han2020multistream} for acoustic modeling, and 4- and 5-gram LMs for 1st pass decoding and rescoring, respectively. This ASR model was trained on approximately 1,900h speech data consisting of in-house phone call data and the Switchboard Cellular Part 1 dataset.

As for the transcript-based sentiment classification component, (either GT or ASR) transcripts were tokenized in a sub-word unit with a max length of 500 tokens. Regarding the classification model, using the SWBD-train and validation sets, we performed a hyperparameter search for finding the optimal weights for the trainable layers 
and also explored different design decisions, like how to utilize the pre-trained BERT embedding (i.e., in the BERT-based system, using the last layer or the last four layers or the sentence embedding provided by BERT -- similar techniques to the ones proposed in \cite{devlin2018bert}). We also tested pre-trained embeddings specifically tuned for sentiment classification (i.e. DistilBERT-SST2), but the plain BERT model was shown to provide better results. The selected baseline model has an embedding dimension of 200, two BLSTM layers with a hidden dimension of 128, trained with a weighted cross-entropy loss as the optimization objective. The selected BERT model uses three BLSTM layers applied on top of the sum of the last four BERT layers and an unweighted cross-entropy loss.

In our experiments, we trained and evaluated systems using both GT and ASR transcripts from the SWBD-sentiments corpus. In this way, we can not only compare our system results with other systems (that usually use GT transcripts), but also measure a performance gap in the case of a real production system that would run on top of ASR results.

\subsection{E2E system experiment setup}

For the E2E speech sentiment analysis systems, we utilized the encoder part of an E2E ASR system that we trained using our in-house data. The ASR model has an encoder-decoder architecture where each component is based on Transformer jointly optimized with the CTC loss~\cite{kim2017joint}. This system included a byte-pair encoding tokenizer (token size of 2,000), and the encoder consisted of 12 transformer blocks that generated a 512 dimension embedding vector. All parameters in the encoder were fixed in the experiments. 

For choosing the pseudo labeler, we evaluated several pre-trained LMs fine-tuned with the Stanford Sentiment Treebank (SST) dataset \cite{socher-etal-2013-recursive}, as shown in Table~\ref{tab:labeler_perform}. In the SST corpus, there are two types of labels, find-grained (5-classes, SST5) and binary (negative/positive, SST2), and we used the SST2 portion to fine-tune the models. The table shows the REC score on the GT transcripts of the evaluation set. Since the family of SST2 models produces the binary classes, we ran this evaluation only on the negative and positive utterances in the evaluation set (thus these numbers are incomparable to the 3-way classification task with the SWBD-Sentiment dataset in our other experiments). Based on the results in the table, we chose BERT-SST2 and XLNet-SST2 as our pseudo labelers.

The sentiment classifier in the E2E system was trained
using the SWBD-train portion of the data.
In the semi-supervised training phase with pseudo labels (Figure~\ref{fig:concept}(b)), we discarded the output layer of the classifier since the pseudo labeler is binary as $\hat{y}=\{Neg, Pos\}$.
Then, we replaced it with a randomly initialized output layer which has a 3 class output in the fine-tuning stage. When fine-tuning the sentiment classifier, we updated the whole parameters in the model.

\begin{table}[t]
\centering
\caption{Pseudo labeler performance on SWBD-holdout for negative and positive classes}
\vspace{-0.3cm}
\label{tab:labeler_perform}
\resizebox{0.7\linewidth}{!}{%
\begin{tabular}{c|cc}
\hlineB{2}
 & Unweighted REC  & Weighted REC \\ \hlineB{2}
BERT-SST2 & 70.99 & 71.05 \\ \hline
DistilBERT-SST2 & 70.95 & 71.16 \\ \hline
RoBERTa-SST2 & 70.16  & 70.25  \\ \hline
XLNet-SST2 & \textbf{72.15} & \textbf{72.19}  \\ \hline \hline
BERT-SST2 & \multirow{2}{*}{69.08} & \multirow{2}{*}{69.20} \\ 
(ASR transcript) & &\\ \hline
\hlineB{2}
\end{tabular}%
}
\vspace{-0.5cm}
\end{table}

\begin{table*}[t]
\centering
\caption{Semi-supervised approach on E2E speech sentiment analysis system evaluation. $\mathcal{S}$:SWBD-train with GT transcripts, $\mathcal{F}$:Fisher with GT transcripts, $\mathcal{S}_{\text{asr}}$:SWBD-train with ASR transcripts, $\mathcal{F}_{\text{asr}}$:Fisher with ASR transcripts. }
\vspace{-0.3cm}
\label{tab:pretrain_full}
\resizebox{0.99\linewidth}{!}{%
\begin{tabular}{c|c|c|ccc|ccc|ccc|ccc}
\hlineB{2}
 & & & \multicolumn{6}{c|}{Validation Set (SWBD-test)} & \multicolumn{6}{c}{Evaluation Set (SWBD-holdout)} \\ \cline{4-15}
Fine-tuning & Pseudo & Semi-supervised  & \multicolumn{3}{c|}{Unweighted} & \multicolumn{3}{c|}{Weighted} & \multicolumn{3}{c|}{Unweighted} & \multicolumn{3}{c}{Weighted} \\ \cline{4-15} 
dataset& labeler & training dataset & REC & PRE & F1 & REC & PRE & F1 & REC & PRE & F1 & REC & PRE & F1 \\ \hlineB{2}
\multirow{5}{*}{SWBD-train (86h)} & - & - & 64.59&	\textbf{68.89}&	\textbf{66.24}&	\textbf{71.41}&	\textbf{70.86}&	\textbf{70.72}&	61.21&	65.92&	62.74&	67.73&	67.89&	66.99 \\ \cline{2-15} 
& BERT-SST2 & $\mathcal{S}$ &63.68 &	67.65 &	65.23 &	70.37 &	69.79 &	69.71 &	62.37 &	66.68 &	63.85 &	68.47 &	68.58 &	67.85 \\ \cline{2-15} 
& BERT-SST2 & $\mathcal{S}$, $\mathcal{F}$ & 64.87&	68.05&	66.15&	70.82&	70.28&	70.31&	63.23&	66.82&	64.55&	\textbf{69.05}&	\textbf{68.77}&	\textbf{68.46} \\ \cline{2-15} 
& XLNet-SST2 & $\mathcal{S}$, $\mathcal{F}$ & 63.64 & 67.56 & 65.17 & 70.45 & 69.86 & 69.78 & 61.61 & 65.48 & 62.94 & 67.73 & 67.58 & 67.06 \\ \cline{2-15} 
& BERT-SST2 & $\mathcal{S}_{\text{asr}}$, $\mathcal{F}_{\text{asr}}$ & \textbf{65.74}&	66.51&	66.11&	70.23&	70.01&	70.10&	\textbf{64.18}&	65.28&	\textbf{64.57}&	68.27&	68.35&	68.14	 \\ \hline \hline
\multirow{5}{*}{SWBD-train (5h)}& - & - & 51.33 & 53.82 & 51.98 & 60.66 & 58.73 & 59.24 & 47.76 & 49.86 & 48.16 & 56.62 & 54.88 & 55.12 \\ \cline{2-15} 
& BERT-SST2 & $\mathcal{S}$ & 54.16 & 58.08 & 54.96 & 62.74 & 61.40 & 61.33 & 52.12 & 56.61 & 53.06 & 60.40 & 59.11 & 58.84  \\ \cline{2-15} 
& BERT-SST2 & $\mathcal{S}$, $\mathcal{F}$ & \textbf{58.72}&	58.67&	\textbf{58.54}&	\textbf{63.92}&	\textbf{63.74}&	\textbf{63.72}&	\textbf{57.45}&	\textbf{57.92}&	\textbf{57.63}&	\textbf{61.98}&	\textbf{61.67}&	\textbf{61.79} \\ \cline{2-15} 
& XLNet-SST2 & $\mathcal{S}$, $\mathcal{F}$ & 58.19 & 57.89 & 58.00 & 62.63 & 63.07 & 62.82 & 56.86 & 57.39 & 56.75 & 60.59 & 61.52 & 60.74 \\ \cline{2-15} 
& BERT-SST2 & $\mathcal{S}_{\text{asr}}$, $\mathcal{F}_{\text{asr}}$ & 54.78&	55.51&	55.02&	61.10&	60.38&	60.67&	52.23&	53.16&	52.60&	57.39&	57.00&	57.10	 \\ \hlineB{2}

\end{tabular}%
}
\vspace{-0.4cm}
\end{table*}

\subsection{Experiment result}

\subsubsection{2-step pipeline}
\vspace{-0.1cm}

Table~\ref{tab:2step_performnance_bert} shows 2-step pipeline experiment results.
First, we observe that the BERT-based models outperform the baseline systems trained with the simple neural net architecture described in Section \ref{sec:2step:describe} across all the metrics. 
Second, the baseline model trained with GT transcripts displayed a remarkable performance drop compared to the baseline model trained with ASR transcripts ($55.55$ vs. $47.66$ Unweighted F1).
On the other hand, the BERT-based model trained on ASR transcripts exhibited similar performance to the one trained on GT transcripts ($64.56$ vs. $63.64$ Unweighted F1).

This table also contrasts performance drop in the baseline and BERT-based systems when using a subset of our training data (5h). The baseline system showed a notable drop of $51\%$ Unweighted F1 score. Although, the BERT-based model showed only a $19\%$ decrease under the same condition. We also observe that the 5h BERT-based model outperforms the baseline model trained on ASR transcripts corresponding to 86 hours by approximately 7\% Unweighted F1 on both the validation and evaluation set (47.38 vs. 50.99, and 51.03 vs. 47.66, respectively). This shows the knowledge embedded in the pre-trained LM can be distilled to the sentiment classifier in the 2-step pipeline even with a small fine-tuning data.

\subsubsection{E2E system}
\vspace{-0.1cm}
Table~\ref{tab:e2e_baseline} shows the performance of E2E systems. Compared to the model presented in \cite{lu2020speech} (which used RNN-T to train an ASR encoder with a word-level token), our joint CTC-Attention model (that uses sub-word unit tokens) showed slightly better performance. We used this CTC-Attention encoder for the rest of the experiments. 

The upper section of Table~\ref{tab:pretrain_full} shows the performances of the proposed semi-supervised training approaches in the E2E system using SWBD-train (86h) and Fisher (2,000h) for pseudo labeling and the full SWBD-train data alone as a fine-tuning set.
We observed a similar performance for all the systems on the validation set, however, the semi-supervised training approaches with the pseudo labeled data generally outperformed the baseline system without the semi-supervised training phase on the evaluation set.
It is also shown that the marginal gain can be obtained when we added the Fisher data on top of SWBD-train as the semi-supervised training dataset.
Comparing the results obtained using XLNet-SST2 and BERT-SST2, we did not see a meaningful difference. 
Besides, using ASR transcripts as input to the pseudo labelers presented similar performance to using GT transcripts, showing the proposed system is robust to ASR errors from a sentiment classification perspective. 

In the bottom section of Table~\ref{tab:pretrain_full}, we show the results of the semi-supervised training approaches with only a 5h subset of SWBD-train for fine-tuning. 
In this setting, the semi-supervised training approaches showed significant improvements as opposed to the baseline without taking an advantage of pseudo labeling.
Also, we observe that the best system showed about 20\% improvement on unweighted F1 score on the evaluation set (57.63\%) compared to the baseline (48.16\%).
Finally, using ASR transcripts for pseudo labeling did not give a similar improvement compared to using GT transcripts, but it shows still a better performance than the baseline.

If we compare to the upper section of the table, that is comparing a fine-tuning set of 86h vs. 5h, we observed that the baseline dropped about 20\% (62.74\% to 48.16\%) while the best system decreased only by 10\% (64.55\% to 57.63\%). 

\begin{figure}[t]
    \vspace{-0.1cm}
    \centering
    \includegraphics[width=0.75\linewidth]{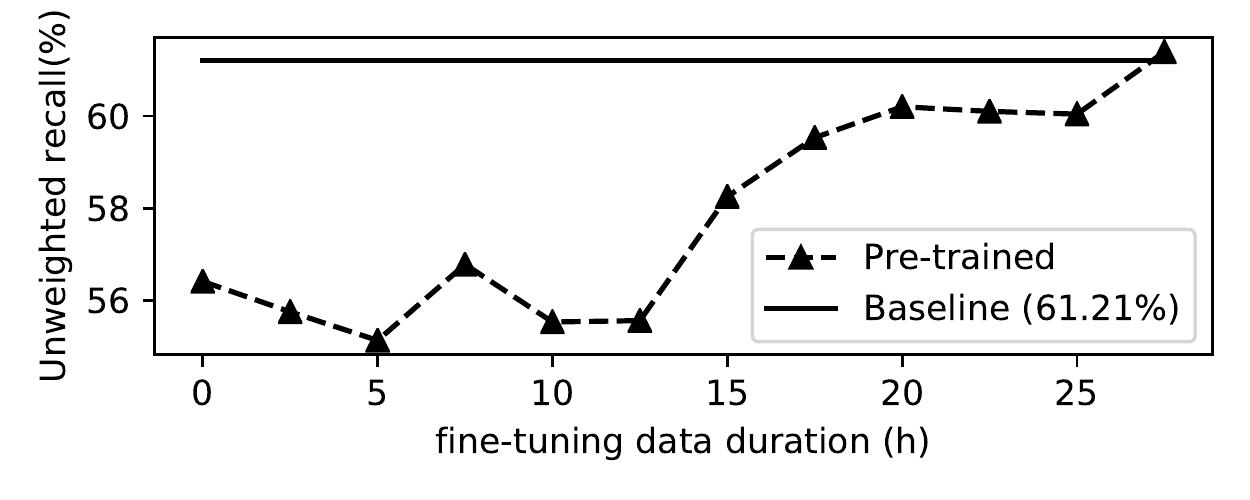}
    \vspace{-0.3cm}
    \caption{Semi-supervised training approach efficiency on evaluation set. Note that baseline used all of SWBD-train set (86h)}
    \vspace{-0.5cm}
    \label{fig:pretrain_efficiency}
\end{figure}

Additionally, we checked the performance by increasing fine-tuning data by 2.5h to predict how much we can reduce the human sentiment annotation job in Figure~\ref{fig:pretrain_efficiency}. From this experiment, we verified the best performing system with a semi-supervised training approach reaches the baseline performance when we use at least a 30h subset of SWBD-train. This means that we can save about 65\% of human sentiment annotation jobs if we use the semi-supervised training approach.

\vspace{-0.1cm}
\subsection{Discussion}
The 2-step pipeline experiments showed that the use of pre-trained LMs can achieve good performance even with a low amount of labeled data. 
Also, the pre-trained LM enables the model to use ASR transcript for training without performance degradation. 
We attribute this robustness to errors in words, to the way BERT is trained -- using token masking techniques.
This result suggests that when new data for speech sentiment analysis is needed, we can skip the expensive human transcripts by using any off-the-shelf ASR.

Given that the most of the SST-2 benchmark results\footnote{https://gluebenchmark.com/} are above 90\% recall using BERT, the text sentiment classifier is not performing well on spoken speech as shown in Table~\ref{tab:labeler_perform}, displaying once more the difficulties of using models trained on written data to adapt to transcribed conversational data. 
However, our noisy pseudo label-based semi-supervised training approach still showed encouraging results, and generally outperformed trained-from-scratch models in various conditions. 
These results suggest that sentiment pseudo labels carry the text-domain sentiment knowledge that could transfer some knowledge to speech sentiment classifiers on a semi-supervised training stage.

A limitation of this study is that we did not consider another text corpus for building the pseudo labeler. 
There are more fine-grained sentiment datasets such as SST-5 (5 classes), IMDb (10 classes), Yelp (5 classes).
We believe these fine-grained data could benefit in different ways to what a binary sentiment system does.
Another limitation is that we did not update the ASR encoder for the speech sentiment analysis system.
We expect that updating the ASR encoder on both semi-supervised training and fine-tuning steps could considerably affect the results.

\section{Conclusion}
In this paper, we investigated an approach to transfer knowledge from the written text to spoken text or speech domain using an LM to reduce and use efficiently the human annotation on speech dataset.
The experiments explored two scenarios, a 2-step pipeline, and an E2E speech sentiment analysis system, to verify the effectiveness of leveraging BERT.
From the experiments, we observed that the proposed approach is able to encode the information robustly and generalize better with less supervision.
While the proposed approaches show improvement in all conditions, we verified that it has a greater advantage in the case where a large amount of audio is available whether it is transcribed or not.

\clearpage
\bibliographystyle{IEEEtran}

\bibliography{paper}

\end{document}